\documentclass{article}

\usepackage{PRIMEarxiv}

\usepackage[utf8]{inputenc} 
\usepackage[T1]{fontenc}    
\usepackage{hyperref}       
\usepackage{url}            
\usepackage{booktabs}       
\usepackage{amsfonts}       
\usepackage{nicefrac}       
\usepackage{microtype}      
\usepackage{lipsum}
\usepackage{fancyhdr}       
\usepackage{graphicx}  

\usepackage{amsmath}   
\usepackage{amssymb}
\graphicspath{{media/}}     

\title{Toward Virtuous Reinforcement Learning: A Critique and Roadmap
\thanks{\textit{AAAI'26: Accepted to the \emph{Machine Ethics From Formal Methods to Emergent Machine Ethics} workshop at the AAAI 2026 Conference.
}} 
}

\author{
  Majid Ghasemi, and Mark Crowley \\
  Electrical \& Computer Engineering \\
  University of Waterloo \\
  \texttt{\{majid.ghasemi, mark.crowley\}@uwaterloo.ca} \\
}

\begin{document}
\maketitle

\begin{abstract}
This paper critiques common patterns in machine ethics for Reinforcement Learning (RL) and argues for a virtue focused alternative. We highlight two recurring limitations in much of the current literature: (i) rule based (deontological) methods that encode duties as constraints or shields often struggle under ambiguity and nonstationarity and do not cultivate lasting habits, and (ii) many reward based approaches, especially single objective RL, implicitly compress diverse moral considerations into a single scalar signal, which can obscure trade offs and invite proxy gaming in practice. We instead treat ethics as policy level dispositions, that is, relatively stable habits that hold up when incentives, partners, or contexts change. This shifts evaluation beyond rule checks or scalar returns toward trait summaries, durability under interventions, and explicit reporting of moral trade offs. Our roadmap combines four components: (1) social learning in multi agent RL to acquire virtue like patterns from imperfect but normatively informed exemplars; (2) multi objective and constrained formulations that preserve value conflicts and incorporate risk aware criteria to guard against harm; (3) affinity-based regularization toward updateable virtue priors that support trait like stability under distribution shift while allowing norms to evolve; and (4) operationalizing diverse ethical traditions as practical control signals, making explicit the value and cultural assumptions that shape ethical RL benchmarks.
\end{abstract}


\section{Introduction}

The growing integration of Artificial Intelligence (AI) systems into mission-critical contexts underscores the need to assess how it is governed ethically and whether its decisions remain sound when ethical tensions arise \cite{vishwanath2023towards}. This attention has helped consolidate the field of Artificial Morality (AM), in which Reinforcement Learning (RL) has been a prominent approach over the past decade \cite{abel2016reinforcement, vishwanath2024reinforcement}. 

Against this backdrop, much of AM literature is based upon existin classical moral theories (deontological, consequentialist, and virtue ethics) to structure RL-based decision-making. In brief, \textbf{deontological ethics} grounds right action in rule or duty compliance, whereas \textbf{consequentialist (utilitarian) ethics} evaluates actions by the value of their outcomes \cite{kant2020groundwork, mill2016utilitarianism}. Operationally, these manifest in RL as constraints, formal specifications, and runtime monitoring on the deontic side \cite{alshiekh2018safe,ayars2016can}, and as reward design, preference learning, and Inverse RL on the consequentialist side \cite{ecoffet2021reinforcement,krening2023q}. 

Most deontic approaches operationalize ethics as rules, constraints, or specifications, which requires principles and inference rules to be encoded in advance and makes systems brittle to ethical uncertainty and novel contexts; such methods also offer compliance guarantees without necessarily cultivating learned dispositions \cite{allen2005artificial,rossi2019building,yu2018building}. On the consequentialist side, reward-centric formulations (including shaping and preference aggregation) risk proxying ethical desiderata into a single objective, inviting perverse incentives and masking trade-offs \cite{wu2018low,abel2016reinforcement}. When multiple moral objectives or theories are considered concurrently, a common cardinal scale is typically absent; consequently, composite rewards become scale-sensitive, and selecting a deployment policy from the Pareto set remains fundamentally underdetermined \cite{ecoffet2021reinforcement,bauer2020virtuous,noothigattu2019teaching}.

By contrast, \textbf{virtue ethics} assigns moral primacy to character dispositions (such as honesty, fairness, and temperance) typically understood as cultivating a balanced ‘mean’ between extremes \cite{vishwanath2023towards,crisp2014aristotle}.
Viewed through an RL lens, virtues are policy-level dispositions acquired through habituation, practice, and socially mediated feedback, whose hallmark is internalization. 
Virtuous behavior persists even when incentives, partners, or contexts shift \cite{bauer2020virtuous,allen2005artificial,vishwanath2023towards,van2011right}. This perspective implies evaluating agents beyond rule compliance or scalar returns, using trait-level summaries and retention under interventions, and keeping ethical trade-offs explicit via multi-objective or orchestration-based treatments rather than collapsing them into a single reward \cite{abel2016reinforcement,noothigattu2019teaching,ecoffet2021reinforcement}.

Accordingly, we do not prescribe a single methodology; rather, we treat virtue ethics as a research frontier for ethical RL and set out a concrete agenda. We (i) shed light on how social learning in RL could be helpful to design virtuous agents, (ii) argue for multi-objective formulations that keep virtue trade-offs explicit instead of collapsing them into a single reward, (iii) examine the potential of affinity-based RL in designing ethical agents, and (iv) suggest a new direction for research via alternative ethical systems from global cultures rather than focusing only on the traditional three main streams of ethics.

\section{Preliminaries}
To set the stage we need to define a few more core concepts for adding complexity to standard single-agent RL.
\subsection{Reinforcement Learning}
We model tasks as Markov decision processes (MDPs) $\mathcal{M}=(\mathcal{S},\mathcal{A},\mathcal{P},r,\gamma)$ with state space $\mathcal{S}$, action space $\mathcal{A}$, transition kernel $P(s' \mid s,a)$, reward $r(s,a)$, and discount factor $\gamma\in[0,1)$. A (possibly stochastic) policy $\pi(a\mid s)$ optimizes the objective \cite{ghasemi2024introduction}
\begin{equation*}
\mathcal{J}(\pi) = \mathbb{E}_{\pi}\!\left[\sum_{t=0}^{\infty}\gamma^{t}\, r(s_t,a_t)\right],
\label{eq:Jpi}
\end{equation*}

where $s_{t+1} \sim \mathcal{P}(\cdot \mid s_t, a_t)$ and $a_t \sim \pi(\cdot \mid s_t)$. 

\subsection{Multi-objective RL}
In standard RL, the objective is scalar: the agent maximizes a single cumulative reward \(r(s,a)\), yielding a clear notion of optimality via the highest expected return \(\mathcal{J}(\pi)\). 
In \textbf{Multi-objective RL (MORL)}, the reward is vector-valued \(\mathbf{r}(s,a)\), so a policy attains a vector of expected returns \(\mathcal{J}(\pi)\) across $m$ distinct goals for $i\in m$.

(To keep ethical trade-offs explicit, we use a vector reward $\mathbf{r}(s,a)\in\mathcal{R}^m$ with objective)
\begin{equation*}
\mathcal{J}(\pi) = \mathbb{E}_{\pi}\!\left[\sum_{t=0}^{\infty}\gamma^{t}\, \mathbf{r}_i(s_t,a_t)\right],
\label{eq:Jcal}
\end{equation*}

A policy $\pi'$ \emph{Pareto-dominates} $\pi$ if $\mathcal{J}_i(\pi') \ge \mathcal{J}_i(\pi)$ for all $i$, with strict inequality for at least one $i$. 
Accordingly, we reason about the Pareto set 
rather than a single scalar summary.
Optimality is therefore characterized by the Pareto front, making MORL better suited to settings where trade-offs (e.g., safety, fairness, efficiency) must remain explicit rather than being collapsed into a single number.




\subsection{Multi-agent extension}
In a 
stochastic game with $\mathcal{N}$ agents, 
$\mathcal{G} = \big(\mathcal{S}, \{\mathcal{A}_i\}_{i=1}^{\mathcal{N}}, \mathcal{P}, \{r_i\}_{i=1}^{\mathcal{N}}, \gamma \big)$, 
a joint policy $\pi = (\pi_1,\ldots,\pi_{\mathcal{N}})$ induces returns $\mathcal{J}_i(\pi)$ for each agent. 
Social objectives can aggregate individual payoffs $\{\mathcal{J}_i\}$, for example via a utilitarian sum, a max--min criterion, 
or a vector-valued multi-objective formulation. 
We will assess robustness to partner and distributional shifts by varying the co-player population and the interaction structure \cite{lee2021joint}.

\section{Directions for Virtuous RL}
In this section, we outline several research directions that, in our view, can move the community closer to developing virtuous agents.

\subsection{Social Learning} Social learning—the acquisition of behavior by observing other agents’ actions and their effects—offers a route to rapid skill acquisition that individual exploration and static struggle to match \cite{bhoopchand2023learning,ye2025efficient}. When integrated with RL (“social RL”), agents can exploit observations of co-present experts or peers to shape their representations and policies without supervised action labels, often yielding faster convergence and stronger generalization for the ego agent \cite{ndousse2021emergent}. Recent work explores this integration across multi-agent settings, including cultural transmission \cite{ye2025efficient}, and cooperation or competition with socially aware objectives \cite{ndousse2021emergent,jaques2019social}. In practice, two complementary methods have been effective: (i) auxiliary predictive objectives that train agents to anticipate how others drive state changes, and (ii) intrinsic social-influence rewards that incentivize generating and leveraging informative signals.

Despite recent advances, the ethical dimensions of social RL remain insufficiently examined. 
Embedding social learning within RL offers a possible path to virtue-centric agents: apprentices can internalize dispositions exhibited by “virtuous” models by observing their actions, the states they seek or avoid, and how they navigate trade-offs, even when optimizing different objectives. 
In multi-agent settings ($\mathcal{J}_i(\pi)$), such dispositions can propagate through cultural transmission and teacher–student interaction, fostering stable, socially desirable behaviors rather than merely maximizing scalar returns. 
Cultural transmission—the domain-general mechanism by which agents acquire and faithfully reproduce knowledge and skills from others—underpins cultural evolution \cite{bhoopchand2023learning}. Hence, having an ethical agent and adding new agents that inherit some behaviors from the ethical one would potentially give us new agents that are ethical to a certain extent all because of culturally transmitted information and behavior.

We treat social RL as the foundation for integrating virtue-sensitive objectives and evaluations. Concretely, we measure the persistence of virtuous dispositions under partner and incentive shifts and test resistance to proxy gaming, thereby shifting the focus from mere rule compliance to cultivated, context-sensitive ethical competence. In addition, we highlight the role of cultural transmission in shaping agents that can be pretrained within an environment and subsequently display ethical behavior in a few-shot setting.
A “virtuous model” does not presuppose an oracle policy that is fully virtuous. Instead, it denotes partial or imperfect sources of normative guidance, such as human demonstrations, curated behavioral datasets, or simplified heuristic teachers. These models are intentionally incomplete: they offer coarse moral direction (for example “avoid clearly harmful actions” or “respect local norms”) but do not solve the task on their own. The role of social learning is to bootstrap from these limited exemplars, allowing the agent to generalize and refine virtues through interaction. The approach is therefore still necessary even when the initial exemplars are not ideal moral agents.


\subsection{Multi-Objective RL}
MORL replaces a single scalar reward  with a vector of objectives and a 
scalarization function, making value trade-offs explicit rather than implicit \cite{deschamps2024multi}. 
This is one way to satisfy one of the requirements for ethical behavior: balancing safety, fairness, efficiency, and other norms instead of over-optimizing one proxy. By assigning each virtue (or moral value) to a reward component, MORL lets us reason over—and learn—policies that respect multiple ethical desiderata without collapsing them into a single number.

MORL provides (i) a representational handle for virtues via vector-valued rewards, 
(ii) optimization tools that preserve trade-offs (Pareto and coverage sets) so agents can remain virtuous under preference or context shifts, and 
(iii) a natural place to incorporate constraints and evaluation (e.g., selecting policies that satisfy minimum thresholds on “virtue” dimensions). 
In short, MORL could be used to turn virtue formation into a learnable, auditable multi-criteria problem that is an actionable direction for building agents whose dispositions remain aligned across partners, incentives, and environments \cite{deschamps2024multi}. This is a direction we suggest that researchers follow and combining this with constrained and risk-aware objectives can be a useful way to incorporate virtue into agents.

\subsection{Affinity-based RL}



In \cite{vishwanath2023exploring}, researchers proposed affinity-based RL (ab-RL) as a way to encode virtues into policies via an interpretable regularization toward a virtue prior $\pi_{0}$. Concretely, they augment the objective with a penalty $\mathcal{L}$ measuring the mean-squared deviation between the learned policy’s action probabilities and $\pi_{0}$, optimizing $\mathcal{J}(\theta)=\mathbb{E}[\mathcal{R}]-\lambda \mathcal{L}$. This yields a tunable mechanism to imprint trait-like tendencies (e.g., an “honest” action) without hard rules or opaque shaping; empirically, in a stochastic role-playing environment inspired by \emph{Papers, Please} game, ab-RL steers behavior toward the virtue prior across arrest-probability regimes, and the learned action frequency $p^{*}(a_{\text{virtue}})$ increases with $\lambda$, approaching $\pi_{0}(a_{\text{virtue}})$ for large regularization. 
Even though this is a major step towards virtuous RL agents, limitations include reliance on a highly simplified single-agent setting with exogenous arrest probabilities, sensitivity to $(\lambda,\pi_{0})$ combinations, and lack of multi-agent social dynamics and Pareto reporting for virtue trade-offs.

From an ethics standpoint, choosing $\pi_{0}$ to represent a virtuous behavioral template turns the regularizer into an explicit mechanism for cultivating stable dispositions rather than merely optimizing a single proxy reward. The prior can encode multi-faceted values (e.g., safety, beneficence, low impact) or be paired with multi-objective rewards, aligning with evidence that ethical alignment is inherently multi-objective and benefits from explicit trade-off handling \cite{deschamps2024multi, vishwanath2023exploring}. However, creating this prior policy can be challenging especially working on problems with higher complexities.
In this way, ab-RL offers a practical, auditable route to train agents whose policies remain close to ethically preferred behavior even as incentives, partners, or contexts shift. This can have substantial impact in designing virtuous agents. However, selecting $(\lambda, \pi_{0})$ can be challenging in higher-complexity settings. We intend to mitigate this with a scheduled $\lambda$ and by regularizing only designated action subspaces, preserving exploration while encouraging trait stability and allowing for a diversity of objectives.

Regularization toward a virtue prior does not assume that the prior itself solves the environment. In practice, virtue priors are intentionally task agnostic and cannot optimize task reward alone. For this reason, we need to treat the regularization weight as a dial that modulates a trade off between task competence and moral conformity. Future work should explicitly characterize this trade off, for instance by quantifying how increases in virtue affinity affect sample efficiency, asymptotic reward, or safe exploration. This reinforces the complementary roles of Social Learning, which learns virtue like features, and ab-RL, which constrains optimization to remain close to them.

\subsection{Broader View of Ethics}

Taking a step back, we believe there is rich potential in global ethical traditions that may be easier to integrate with RL than the usual triad (consequentialist, deontological, virtue). A broader machine ethics agenda should ask: 
How can agents learn context-sensitive role obligations and restorative norms inspired by Confucian \textit{Ren} \cite{li2003core}, emphasizing holistic growth rather than outcome maximization? 

When should systems favor minimal, non-coercive interventions in the Daoist spirit of \textit{Wu-Wei} \cite{van2025daoist}, rewarding low-impact, reversible actions that still meet goals? From Persian Akhlaq, Adab, and the Maqasid \cite{nasr2012anthology}, which dispositions (truthfulness, generosity, restraint) should persist as policy traits, and how do we test whether actions advance public benefit, justice, and dignity rather than proxies?

These notions are not universal primitives. Concepts such as public benefit, justice, and dignity have culturally specific interpretations and can vary across communities. Our point is not to fix these concepts once and for all, but to argue that RL benchmarks and evaluation protocols should make the underlying value choices explicit instead of assuming universal agreement.


\subsection{The Proposed Architecture}
\begin{figure}[t]
  \centering
  \includegraphics[width=0.98\linewidth]{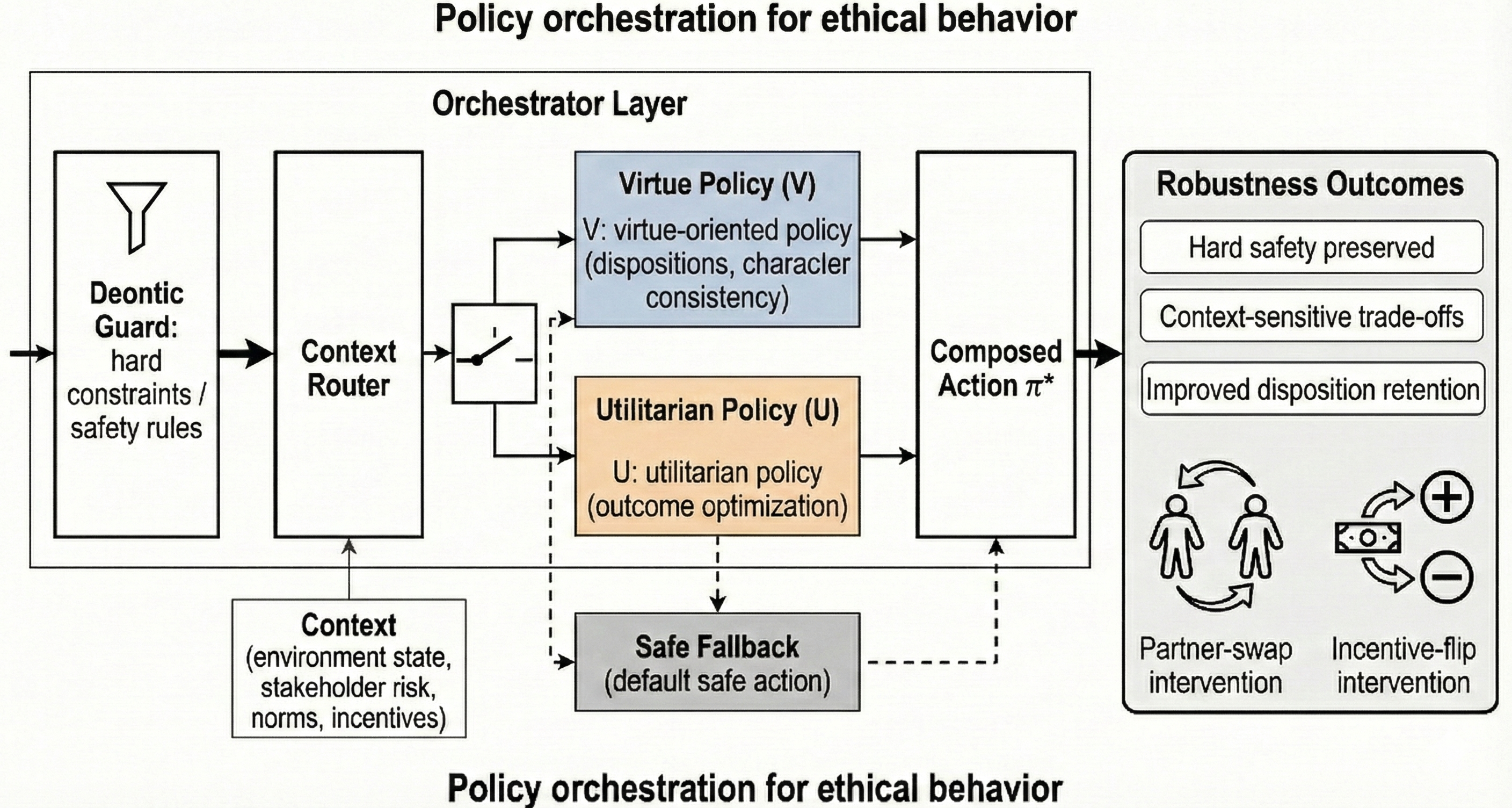}
  \caption{An orchestrator layer first enforces non-negotiable constraints (deontic guard), then selects between a virtue-oriented policy (V) and a utilitarian policy (U) based on context, with a safe fallback. This composition preserves hard safety while enabling context-sensitive trade-offs, improving \emph{disposition retention} under partner-swap and incentive-flip interventions.}
  \label{Fig:Fig_1}
\end{figure}

We propose a modular \emph{Orchestrator Layer} (Figure \ref{Fig:Fig_1}) designed to address the inherent tension between rigid safety and flexible ethical reasoning. This architecture is presented not as a final solution, but as a conceptual framework for how hybrid ethical systems might be structured. 

The pipeline begins with a \textbf{Deontic Guard}, which acts as a hard filter for non-negotiable safety constraints. Once safety is guaranteed, a \textbf{Context Router} determines the appropriate ethical modality. This raises an open question: can a system effectively switch between a \textbf{Virtue Policy ($V$)}, prioritizing character consistency, and a \textbf{Utilitarian Policy ($U$)}, which optimizes for external outcomes? 

Central to our proposal is the role of \textbf{Social RL} in operationalizing the Context Router. We hypothesize that Social RL is essential for learning the subtle "switching" logic required to navigate fluid social norms. By treating the selection between $V$ and $U$ as a high-level policy decision learned through multi-agent interaction, the system can optimize for long-term social cooperation rather than immediate rewards. 

Furthermore, we suggest that Social RL enables superior \textbf{disposition retention}. By training the router to value social reputation and consistency across diverse interactions, the agent may resist "policy drift" during \emph{incentive-flip} or \emph{partner-swap} interventions. This architectural composition aims to ensure that the final action ($\pi^*$) remains ethically robust even when environmental incentives are volatile.


\subsection{Final Remarks}
To summarize we are advocating broadening the design space for ethical agents beyond single-paradigm solutions. In particular, focusing on machine ethics within RL is promising because RL affords controllable, task-specific environments and systematic evaluation. Crucially, there is no one-size-fits-all resolution to ethical dilemmas: a given environment may require multiple ethical perspectives to be considered and combined (Figure~\ref{Fig:Fig_1}). Concretely, the same agent could acquire complementary sub-policies across distinct sub-environments. For example, (i) a virtue-oriented policy learned via social RL, (ii) a utilitarian policy optimized for aggregate outcomes, and (iii) a deontic policy constrained by explicit rules. When similar states arise at deployment, an \textit{orchestrator agent} could select, or blend, the relevant sub-policy, yielding an overall policy that is ethical yet context-sensitive. Modular RL \cite{simpkins2019composable}, may provide a natural scaffold for such acquisition and composition.

An additional dimension to ethical machine reasoning comes from the need to deal with and learn about soft concepts while simultaneously being to rigorously enforce well established ethical and functional constraints and rules. Here \emph{Formal methods} provide a perfect complement to machine learning. 
Temporal-logic specifications and automata-based synthesis enable stating non-negotiable constraints and deriving policies that satisfy them by construction, while model checking verifies compliance in the induced MDP \cite{dennis2016formal, camacho2019ltl}. 
When full synthesis is infeasible, runtime monitoring and shielding offer verifiable safety envelopes around RL during exploration and deployment \cite{alshiekh2018safe}. Structured task formalisms such as reward machines \cite{icarte2018using} improve sample efficiency and track progress toward ethical requirements, and contract-based design supports compositional guarantees in multi-agent settings, whether it comes to designing virtuous agents or other types.

A more fundamental question is whether we can design agents that rely on social RL, affinity based methods, and other ethical mechanisms without ultimately reducing all ethical guidance back into a single scalar reward. If this reduction is unavoidable, it may reflect a structural limitation of standard RL rather than only a design choice in particular algorithms.


\section{Conclusion}


This paper argued that prevailing deontic and reward centric approaches to ethical RL face structural limits: rule based methods are brittle under ambiguity and nonstationarity, and scalar rewards often compress plural values into a single objective that invites proxy gaming and hides trade offs. We sketched a virtue centric alternative that treats ethical behavior as learned dispositions that remain stable as incentives, partners, and context change, rather than as simple rule compliance or return maximization. The agenda combines four complementary elements: social learning in multi agent settings to acquire virtue like patterns from imperfect exemplars; multi objective and constrained formulations that keep value conflicts explicit and incorporate risk sensitive bounds on harm; affinity based regularization toward modifiable virtue priors that support trait like robustness while tracking evolving norms; and the operationalization of broader ethical traditions as implementable control mechanisms, such as role sensitive policies, low impact penalties, and repair oriented interventions. This roadmap is not a complete blueprint but a starting point that highlights open problems in constructing and revising virtue priors, handling cross cultural variation in norms, and measuring moral task trade offs in RL benchmarks.

\bibliographystyle{unsrt}  
\bibliography{references}

\end{document}